 \documentclass[pmlr,twocolumn,10pt]{jmlr} % W&CP article

\usepackage{graphicx}
\usepackage{booktabs}
\usepackage{siunitx}

\makeatletter
\newcommand*{\@titlefoot}{\scriptsize\copyright\space\@jmlryear}
\makeatother

\usepackage[switch]{lineno}

\usepackage[table]{xcolor}
\usepackage{multirow}

\theorembodyfont{\upshape}
\theoremheaderfont{\scshape}
\theorempostheader{:}
\theoremsep{\newline}

\title[Leveraging Long Context in Biomedical Vision–Language Models]{No Tokens Wasted: Leveraging Long Context in Biomedical Vision–Language Models}

\usepackage{hyperref}

\author{%
\Name{Min Woo Sun}$^{1*}$\Email{minwoos@stanford.edu}\\
\Name{Alejandro Lozano}$^{1*}$ \Email{lozanoe@stanford.edu} \\ % \footnotemark[1] \Email{lozanoe@stanford.edu}\\
\Name{Javier Gamazo Tejero}${^2}$ \Email{javierg@nvidia.com}\\
\Name{Vishwesh Nath}${^2}$ \Email{vnath@nvidia.com}\\
\Name{Xiao Xiao Sun}$^{1}$ \Email{xxsun@stanford.edu}\\
\Name{James Burgess}$^{1}$ \Email{jmhb@stanford.edu}\\
\Name{Yuhui Zhang}$^{1}$ \Email{yuhuiz@stanford.edu}\\
\Name{Kun Yuan}$^{1}$ \Email{kun@unistra.fr}\\
\Name{Robert Tibshirani}$^{1}$ \Email{tibs@stanford.edu}\\
\Name{Sean Huver}${^2}$ \Email{shuver@nvidia.com}\\
\Name{Serena Yeung-Levy}$^{1}$ \Email{syyeung@stanford.edu}\\
\addr $^{*}$Equal contribution \\
\addr $^{1}$Stanford University, USA \\
\addr $^{2}$NVIDIA, USA
}

\makeatletter
\providecommand{\@jmlrproceedings}{}
\makeatother

\begin{document}

\maketitle

\begin{abstract}

Embedding vision–language models (VLMs) are typically pretrained with short text windows ($<$77 tokens), which forces the truncation of long-format captions. Yet, the distribution of biomedical captions from large-scale open source literature reveals that a huge portion of captions far exceed 77 tokens. To this end, we investigate the impact of pretraining on long-format biomedical captions by extending the context length of text encoders in VLMs. We find that longer context (thus, enabling additional supervision provided in long-format captions) correlates with better retrieval and classification performance. Given this finding, we introduce BIOMEDICA-LongCAP, a dataset of 1M image–caption pairs enriched with context-aware descriptions from full-text articles, providing longer and additional textual supervision. Using BIOMEDICA-LongCAP, we train BMC-LongCLIP, a long-context biomedical VLM with a text encoder supporting windows of up to 512 tokens. 
Our model extends context capacity by 6.6×, reducing token waste from 55\% to just 2.2\%. On long-caption retrieval benchmarks, BMC-LongCLIP achieves up to +30\% absolute gains in Recall@1 and +2\% average improvements in classification, while also converging faster than short-context. Our results demonstrate that long-context modeling is a promising direction for advancing biomedical VLMs.

% Our model increases context capacity by 6.6×, yielding a +4.8\% gain in zero-shot accuracy and up to +9.4\% Recall@1 on long-caption and image retrieval. 

% Our model extends context capacity by 6.6×, reducing token waste from 55\% to just 2.2\%. On long-caption retrieval benchmarks, BMC-LongCLIP achieves up to 30 point absolute gains in Recall@1, while also delivering improvements in classification accuracy (+2 points on average). Our results demonstrate that long-context modeling is a promising direction for advancing biomedical VLMs.

% Our model increases context capacity by 6.6×, yielding a +4.8\% gain in zero-shot accuracy and up to +9.4\% Recall@1 on long-caption and image retrieval. Our results demonstrate that long-context modeling is a promising direction for advancing biomedical VLMs.

\end{abstract}

% Extending context length accelerates convergence and yields large relative gains: on CXR retrieval, Recall@10 improves by over 2× compared to baselines; on PMC, BMC-LongCLIP achieves up to 90\% relative gains over MedSigLIP and slightly surpasses BiomedCLIP. For zero-shot classification across 6 biomedical domains, BMC-LongCLIP delivers a +20\% relative improvement over BiomedCLIP.

% Extending context length accelerates convergence and yields large relative gains: on CXR retrieval, Recall@10 improves by over 2× compared to prior baselines; on PMC, BMC-LongCLIP achieves up to 90\% relative gains over MedSigLIP and slightly surpasses BiomedCLIP. 

% For zero-shot classification across six biomedical domains, BMC-LongCLIP delivers a +20\% relative improvement over BiomedCLIP.
\paragraph*{Data and Code Availability}
% This initial paragraph is \textbf{mandatory}. Briefly state what data you
% use (including citations if appropriate) and whether and where the data are
% available to other researchers.
% If you are not sharing code, you must explicitly state that you are not
% making your code available. If you are making your code available, then
% at the time of submission for review, please include your code as
% supplemental material or as a code repository link; in either case, your
% code must be anonymized. If your paper is accepted, then you should
% de-anonymize your code for the camera-ready version of the paper. \emph{If
% you do not include this data and code availability statement for your
% paper, or you provide code that is not anonymized at the time of
% submission, then your paper will be desk-rejected.} Your experiments later
% could refer to this initial data and code availability statement if it is
% helpful (e.g., to avoid restating what data you use).

% For CLIP model training, we use the publicly available OpenCLIP repository.\footnote{\url{https://github.com/mlfoundations/open_clip}}  We adapt OpenCLIP and add model config for BMC-longCLIP.\footnote{\url{https://github.com/minwoosun/open_clip}}
% We train on the BIOMEDICA dataset~\cite{lozano2025biomedica}, which is openly available on Hugging Face.\footnote{\url{https://huggingface.co/datasets/BIOMEDICA}} 
% We additionally use MIMIC-CXR~\cite{mimic} for evaluation, which is available through PhysioNet under credentialed access. Model weights will be made available on Hugging Face \footnote{\url{https://huggingface.co/BIOMEDICA/models/BMC-LongCLIP}.

For CLIP training, we adapt OpenCLIP using our forked version with BMC-LongCLIP config\footnote{\url{https://github.com/minwoosun/open_clip_bmc}}. 
We train on the BIOMEDICA dataset~\cite{lozano2025biomedica} and additionally use MIMIC-CXR~\cite{mimic} 
for evaluation (credentialed access via PhysioNet). Model weights will be released on Hugging Face \footnote{\url{https://huggingface.co/BIOMEDICA/models/BMC-LongCLIP}}.

% \paragraph*{Institutional Review Board (IRB)}
% This research does not require IRB approval.

\section{Introduction}
\label{sec:intro}

%GMAI models will be capable of carrying out a diverse set of tasks using very little or no task-specific labelled data.

%Multimodal foundation models (FMs) hold immense potential to to transform  medical and biological applications \cite{moor2023foundation}, from daignostcs to prognostic and even new doicoveries

\begin{figure*}[htbp]
  \centering
 % \fbox{\rule{0pt}{200pt}\rule{150pt}{0pt}} % height=100pt, width=150pt
  \includegraphics[width=0.8\linewidth]{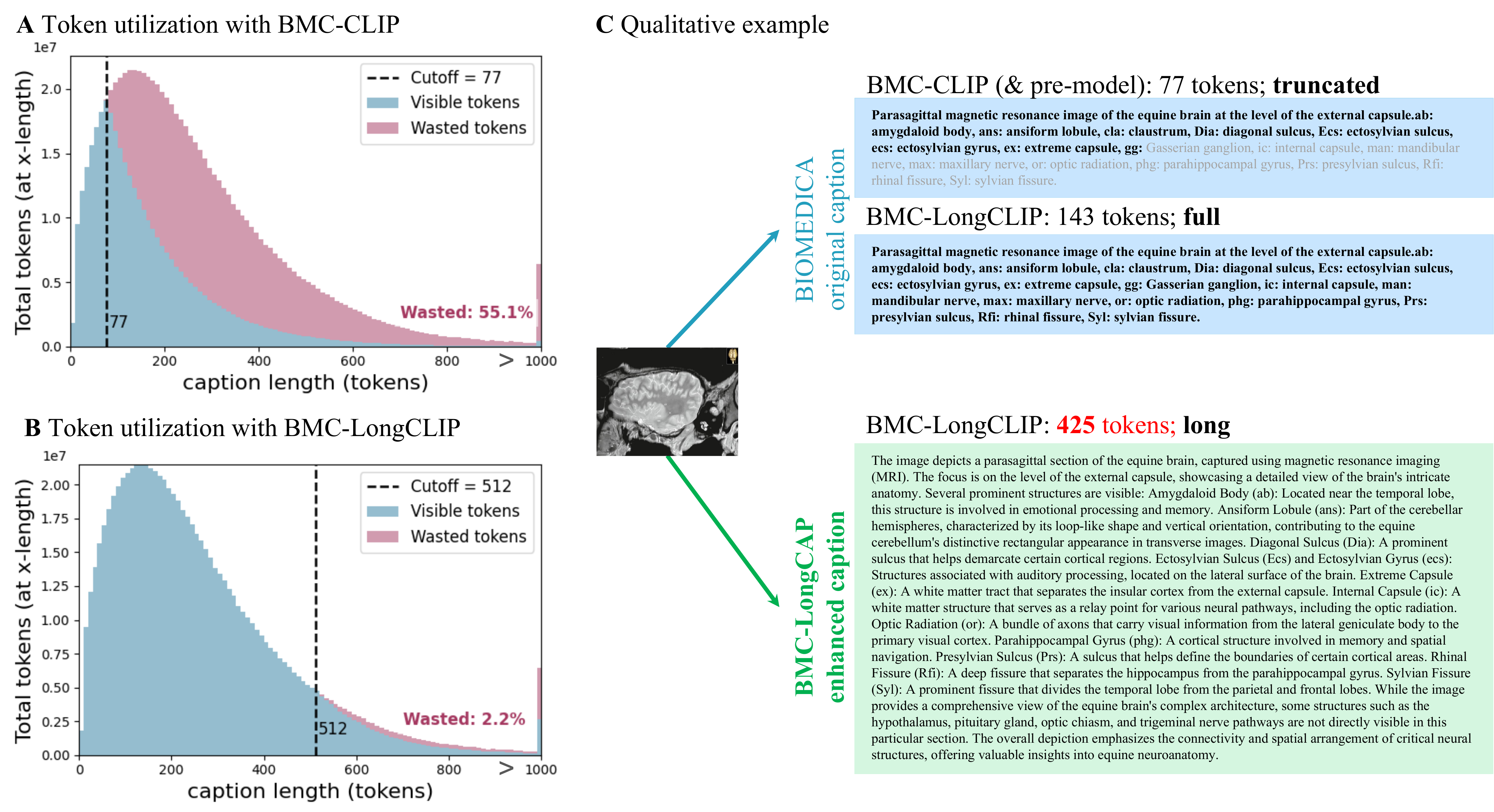}
  \caption{
  % (A) BIOMEDICA-6M caption token length distribution 
  (A) Distribution of BIOMEDICA-6M caption token usage with a cutoff of 77 tokens. The blue histogram represents tokens visible to the model, while the pink histogram represents wasted tokens truncated beyond the cutoff (corresponding to 434 million tokens or 55\% of total tokens ). (B) Distribution with a cutoff of 512 tokens, showing substantially reduced token waste of 2.2\% (17M tokens). 
  (C) Qualitative examples of BIOMEDICA-6M and BIOMEDICA-LongCAP captions, showing truncated vs. full captions, as well as our enhanced captions.}
  \label{fig: motivation-and-example}
\end{figure*}

Multimodal foundation models hold immense potential to advance medical practice and biological science \cite{moor2023foundation}. In particular, transformer-based architectures trained on large image–text datasets have set state-of-the-art performance in tasks such as zero-shot image classification and cross-modal retrieval. Despite these advances, a key limitation remains: current multimodal embedding models (e.g., CLIP \cite{radford2021learning}) are trained using a restricted text context length---typically capped at 77 tokens---which is often insufficient to capture the rich semantics and complexity of high-throughput biomedical images \cite{zhang2024long}.  As a result, it is common practice to truncate long-form textual descriptions during training and inference, discarding valuable information. For example, as shown in Figure~\ref{fig: motivation-and-example},  at a 77 token cutoff,  more than 434 million tokens are not used when pretraining with the BIOMEDICA dataset \cite{lozano2025biomedica} (the largest biomedical image caption dataset). 

Beyond architectural limitations, capturing the semantics of biomedical images through text remains a major bottleneck. Prior work has leveraged open-access scientific articles to curate large collections of image–caption pairs \cite{zhang2023biomedclip, lozano2025biomedica}; however, these captions often fail to fully convey the visual content present in an image. For instance, critical descriptive details are frequently embedded in inline references within the corresponding scientific manuscript (such as the analysis of a figure) and omitted from the corresponding image captions.

\begin{figure*}[htbp]
  \centering
 % \fbox{\rule{0pt}{200pt}\rule{150pt}{0pt}} % height=100pt, width=150pt
  \includegraphics[width=1.0\linewidth]{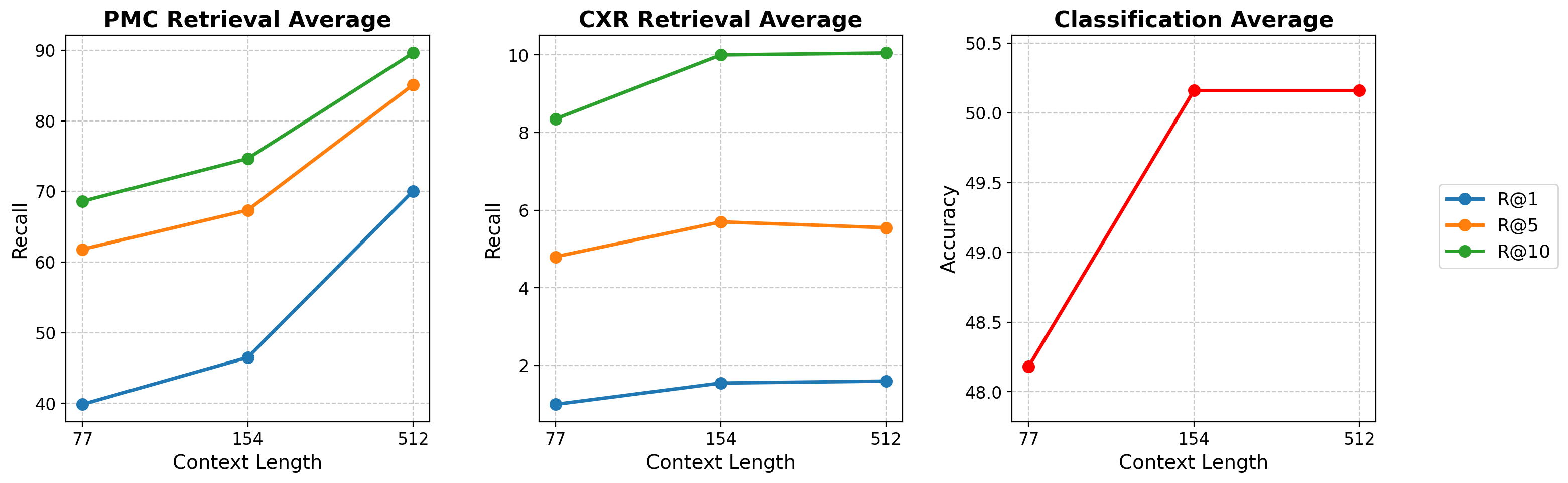}
  \caption{\textit{Context-length ablation results of BMC-LongCLIP trained with 77, 154, and 512 tokens.} (Left) Average retrieval performance (Recall@K) on the PMC long-caption benchmark. (Middle) Average retrieval performance on the CXR benchmark. (Right) Average zero-shot classification accuracy across biomedical datasets. Longer context improves retrieval and classification, with the largest gains on PMC.}
  \label{fig: performance}
\end{figure*}

Given these challenges, the impact of pretraining multimodal embedding models with highly descriptive image captions remains largely unexplored. In this work, we investigate the effects of pretraining  biomedical multimodal embedding models with long-captions by introducing the following contributions:

\begin{itemize}

\item \textbf{BIOMEDICA-LongCAP}: We present a dataset of 1M biomedical image–caption pairs, with captions enriched through LLM-based augmentation leveraging contextual information from the corresponding source text.

\item \textbf{BMC-LongCLIP}: We pretrain CLIP on BIOMEDICA and BIOMEDICA-LongCAP using context lengths of 77, 154, and 512 tokens to study how scaling text context length (thus reducing token waste) impacts model convergence and downstream zero-shot performance.

\item \textbf{Multimodal Long-Text  Bench}: We introduce two novel biomedical benchmarks designed to evaluate long-text multimodal retrieval.

\end{itemize}

Our empirical findings show that (1) pretraining CLIP with longer context lengths accelerates convergence, (2) improves zero-shot classification performance on short captions, and (3) unlocks real-world long-context retrieval applications. 

By extending the text encoder’s context window by 6.6×, our model reduces token waste from 55\% to 2.2\%, enabling substantially more supervision from long biomedical captions. On long-caption retrieval benchmarks, BMC-LongCLIP achieves up to 30 point absolute gains in Recall@1, while also delivering 2 point average improvements in classification accuracy. The model also converges faster than short-context baselines, demonstrating the efficiency benefits of longer context windows during training. These findings highlight long-context modeling as a promising direction for advancing biomedical vision–language models.

% Our best model extends context length 6.6× beyond the previous state of the art while maintaining 4.8\% and up to 9.4\% average gains in zero-shot classification and in Recall@1 for long-caption image retrieval.

%swe release our models and datasets to foster further advances in long-context  multi-modal learning

% \noindent

% \begin{figure}[htbp]
% \floatconts
%   {fig:subfigex2}
%   {\caption{Another Example With Sub-Figures.}}
%   {%
%     \subfigure[A Small Circle][c]{\label{fig:circle2}%
%       \includegraphics[width=0.1\linewidth]{images/circle}}%
%     \qquad
%     \subfigure[A Square][c]{\label{fig:square2}%
%       \includegraphics[width=0.2\linewidth]{images/square}}
%   }
% \end{figure}

% \section{Related Work}
\section{Methods}
\label{sec:methods}

\textbf{Datasets}: 
We pretrain all models on the 6M biomedical image–caption subset of the BIOMEDICA-24M dataset \cite{lozano2025biomedica}. 
In addition, we construct a derived dataset, \textbf{BIOMEDICA-LongCAP}, consisting of 1M image–caption pairs. 
Each LongCAP caption is created by enriching the original figure caption with contextual information from the corresponding article (e.g., in-line mentions, abstract text, and acronym expansions).
A VLM-based augmentation pipeline then refines these captions to retain only features that are visually supported by the image (see Appendix~\ref{apd:longcap} for details). 
We use BIOMEDICA-6M for all baseline pretraining, and BIOMEDICA-LongCAP specifically for the BMC-LongCLIP+ variant. The average caption token length is 127 for BIOMEDICA-6M and 323 for BIOMEDICA-LongCAP.

% We used the  6M biomedical  image–caption subset from the BIOMEDICA 24M \cite{lozano2025biomedica} dataset to pretrain all our models. Furthermore,  we built BIOMEDICA-LongCAP, a set of 1M image–caption pairs generated
% via captions enrichment through  a LLM-based augmentation  pipeline. To this end, captions were expanded with contextual information grounded on corresponding source text such as in-line mentions and refined to retain  image-supported features (see Appendix section \ref{} for details).

% \textbf{Modeling}: We introduce \textbf{BMC-LongCLIP}, a long-context biomedical vision–language model designed to align images with long captions. 
% The model combines a ViT-L/14 CLIP vision encoder (304M) pretrained on DFN-2B \cite{DFN} with BioClinical ModernBERT (150M) \cite{bioclinicalmodernbert}, a long-context text encoder trained on 53.5B biomedical tokens with an 8,192-token window. 

\textbf{Modeling}: We introduce \textbf{BMC-LongCLIP}, a long-context biomedical VLM designed to align images with extended text descriptions. The model pairs a ViT-L/14 CLIP vision encoder (304M) pretrained on DFN-2B \cite{DFN} with BioClinical ModernBERT (150M) \cite{bioclinicalmodernbert}, a long-context text encoder pretrained on 53.5B biomedical tokens with an 8,192-token context window.

% To process the long captions in BIOMEDICA-LongCAP, we introduce BMC-LongCLIP, which pairs a ViT-L/14 CLIP vision encoder pretrained on DFN-2B \cite{DFN} with BioClinical ModernBERT \cite{bioclinicalmodernbert}, a text encoder trained on 53.5B biomedical tokens with an 8,192-token context window.

\subsection{Benchmarks}
We build and evaluate on two complementary benchmarks that stress different aspects of long-context retrieval.   
% We created and collected the following benchmarks:

\noindent \textbf{MIMIC-CXR radiology report (CXR)}
We constructed a long-text benchmark from the MIMIC-CXR dataset \cite{mimic} by pairing chest X-ray images with their full radiology reports. We sampled 1,000 unique image–report pairs, where the reports provide free-text descriptions.

\noindent \textbf{PubMed Long-Caption (PMC)}
From 1,000 PMC-OA articles (restricted to recent 2025 publications), we construct long captions by concatenating inline references with figure captions, testing retrieval in scientific literature where extended technical context is essential.

\begin{table*}[t]
\centering
\scriptsize
\setlength{\tabcolsep}{3pt} % restore some padding
\renewcommand{\arraystretch}{0.85} % tighter rows
\caption{Text$\rightarrow$Image (T2I) and Image$\rightarrow$Text (I2T) retrieval on long-text CXR and PMC benchmarks, reported as Recall@K (higher is better; \textbf{bold} = best, \underline{underline} = second-best). \textit{Panel A} shows the context-length ablation for BMC-LongCLIP; \textit{Panel B} benchmarks against prior models.}
\label{tab:retrieval_all}
\begin{tabular}{l l l l
*{3}{S[table-format=2.1, table-number-alignment=center]}
*{3}{S[table-format=2.1, table-number-alignment=center]}}
\toprule
\textbf{Benchmark} & \multicolumn{3}{c}{\textbf{Model}} &
\multicolumn{3}{c}{\textbf{T2I}} & \multicolumn{3}{c}{\textbf{I2T}} \\
\cmidrule(lr){2-4}\cmidrule(lr){5-7}\cmidrule(lr){8-10}
 & \textbf{Name} & \textbf{Context} & \textbf{Batch}
 & \textbf{R@1} & \textbf{R@5} & \textbf{R@10}
 & \textbf{R@1} & \textbf{R@5} & \textbf{R@10} \\
\midrule
\multicolumn{10}{l}{\textit{Panel A: Context-length ablation}} \\
 & BMC-LongCLIP  & 77 & 8K & 1.3 & 4.7 & 9.4  & 0.7 & 4.9 & 7.3 \\
CXR & BMC-LongCLIP & 154 & 8K & \underline{1.7} & \textbf{5.9} & \textbf{10.7} & \textbf{1.4} & \textbf{5.5} & \underline{9.3} \\
 & BMC-LongCLIP & 512 & 8K & \textbf{1.8} & \underline{5.6} & \underline{10.3} & \textbf{1.4} & \textbf{5.5} & \textbf{9.8} \\
\cmidrule(lr){2-10}
 & BMC-LongCLIP  & 77 & 8K & 37.2 & 59.8 & 66.4 & 42.5 & 63.8 & 70.8 \\
PMC & BMC-LongCLIP & 154 & 8K & \underline{44.2} & \underline{64.9} & \underline{72.5} & \underline{48.8} & \underline{69.8} & \underline{76.8} \\
 & BMC-LongCLIP & 512 & 8K & \textbf{68.9} & \textbf{84.3} & \textbf{89.3} & \textbf{71.2} & \textbf{85.9} & \textbf{89.9} \\
\midrule
\multicolumn{10}{l}{\textit{Panel B: Baseline comparison}} \\
 & PMC-CLIP            & 77 & 128 & {0.0} & {0.5} & {0.7} & {0.2} & {1.0} & {1.6} \\
 & BiomedCLIP          & 256 & 4K & 0.5 & 2.6 & 5.7  & 0.6 & 3.3 & 5.5 \\
CXR & BMC-CLIP            & 77 & 8K & 0.1 & 1.1 & 2.9  & 0.3 & 1.9 & 3.4 \\
 & BMC-LongCLIP    & 512 & 8K & 1.8 & 5.6 & 10.3  & 1.4 & 5.5 & 9.8 \\
 & BMC-LongCLIP  & 512 & 16K & \textbf{2.1} & \textbf{9.5} & \underline{12.1} & \underline{2.5} & \underline{9.1} & \underline{14.2} \\
 & BMC-LongCLIP+  & 512 & 16K & \underline{1.9} & \underline{7.1} & \textbf{12.2} & \textbf{3.0} & \textbf{9.5} & \textbf{14.5} \\
\cmidrule(lr){2-10}
 & PMC-CLIP            & 77 & 128 & 0.2 & 0.7 & 1.2 & 0.1 & 0.7 & 1.2 \\
 & MedSigLIP           & 77 & N/A & 20.1 & 37.0 & 46.0 & 30.9 & 49.0 & 60.1 \\
 & BiomedCLIP          & 256 & 4K & 68.8 & 86.2 & 91.1 & 73.3 & 89.3 & \underline{93.7} \\
PMC & BMC-CLIP    & 77 & 8K & {49.0} & {67.6} & {74.0} & {40.8} & {60.4} & {68.4} \\
 & BMC-LongCLIP    & 512 & 8K & 68.9 & 84.3 & 89.3  & 71.2 & 85.9 & 89.9 \\
 & BMC-LongCLIP   & 512 & 16K & \underline{80.0} & \textbf{92.3} & \textbf{95.1} & \textbf{80.8} & \textbf{91.2} & 93.5 \\
 & BMC-LongCLIP+  & 512 & 16K & \textbf{80.8} & \underline{91.2} & \underline{94.4} & \underline{79.7} & \underline{90.6} & \textbf{93.8} \\
\bottomrule
\end{tabular}
\end{table*}

\noindent \textbf{Zero-shot Classification.}  
For zero-shot image classification, we evaluate on 39 benchmarks spanning biology, radiology, dermatology, and pathology, as collected and described in \cite{biomedica} (see Appendix section \ref{apd:zeroshot} for more details).

\subsection{Baselines} 
% small table: model name and context length ?
To contextualize our results, we benchmark our models  against several  baselines, including: PMC-CLIP    \cite{eslami2023pubmedclip}, BiomedCLIP \cite{zhang2023biomedclip}, MedSigLIP \cite{sellergren2025medgemma}, and BMC-CLIP  \cite{lozano2025biomedica},

\section{Experiments}
\label{sec:experiments}

\subsection{Context-length ablation}
We assess the impact of extending text context length, thus reducing token waste on downstream zero-shot performance. To this end, models were trained with context windows of 77, 154, and 512 tokens under identical settings (batch size, learning rate, optimizer, epochs; as described in appendix section \ref{apd:param}).

\subsection{Batch size and BIOMEDICA-LongCAP}
We investigate the effect of scaling training batch size while holding other settings fixed, comparing 8K and 16K global batches. In addition, we trained BMC-LongCLIP on both BIOMEDICA-6M and BIOMEDICA-LongCAP, a 1M image–caption dataset with captions enriched from full-text context, using a 16K batch size and 512-token context window. We denote this model as BMC-LongCLIP+.

\begin{table*}[t]
\centering
\scriptsize
\setlength{\tabcolsep}{3pt} % keep it compact
\renewcommand{\arraystretch}{0.9} % tighter rows
\caption{\vspace{-0.6em}
Zero-shot classification results of different vision–language models across six biomedical domains. Numbers report average accuracy per domain (higher is better; \textbf{bold} = best, \underline{underline} = second-best). \textit{Panel A} shows ablations of BMC-LongCLIP; \textit{Panel B} benchmarks against prior models.}
\label{tab:zeroshot}
\vspace{-0.6em}
\begin{tabular}{l c c
*{6}{S[table-format=2.2]}S[table-format=2.2]}
\toprule
\multicolumn{3}{c}{\textbf{Model}} &
{\textbf{Biology}} & {\textbf{Dermatology}} & {\textbf{Microscopy}} &
{\textbf{Ophthalmology}} & {\textbf{Pathology}} & {\textbf{Radiology}} & {\textbf{Avg}} \\
\cmidrule(lr){1-3}
\textbf{Name} & \textbf{Context} & \textbf{Batch} &
 &  &  &  &  &  & \\
\midrule
\multicolumn{10}{l}{\textit{Panel A: Context-length ablation}} \\
BMC-LongCLIP    & 77  & 8K   & \textbf{40.82} & 40.69 & 46.04 & \textbf{59.80} & 42.28 & 59.42 & 48.18 \\
BMC-LongCLIP    & 154 & 8K   & \underline{37.21} & \underline{51.34} & \textbf{55.76} & 49.88 & \textbf{47.32} & \underline{59.47} & \textbf{50.16} \\
BMC-LongCLIP   & 512 & 8K   & 34.95 & \textbf{55.16} & \underline{53.37} & \underline{55.41} & \underline{42.87} & \textbf{63.20} & \textbf{50.16} \\
\midrule
\multicolumn{10}{l}{\textit{Panel B: Baseline comparison}} \\
PMC-CLIP        & 77  & 128  &  7.75 & 12.59 & 10.91 & 23.26 & 19.11 & 38.64 & 18.71 \\
MedSigLIP       & 77  & N/A  & 33.98 & 20.13 & 34.56 & 38.23 & 39.74 & 53.03 & 36.61 \\
BiomedCLIP      & 256 & 4K   & 34.07 & 36.01 & 49.71 & 37.36 & 38.40 & 56.05 & 41.93 \\
BMC-CLIP        & 77  & 8K   & 34.08 & \textbf{65.81} & \underline{50.09} & 36.74 & 41.21 & 59.15 & 47.85 \\
BMC-LongCLIP    & 512 & 8K   & \underline{34.95} & 55.16 & \textbf{53.37} & \textbf{55.41} & 42.87 & \underline{63.20} & \textbf{50.16} \\
BMC-LongCLIP    & 512 & 16K  & \textbf{34.98} & 38.80 & 23.16 & 48.79 & \underline{46.25} & 52.79 & 40.79 \\
BMC-LongCLIP+   & 512 & 16K  & 34.34 & \underline{55.54} & 37.30 & \underline{53.05} & \textbf{47.65} & \textbf{66.99} & \underline{49.48} \\
\bottomrule
\end{tabular}
\vspace{-0.6em}
\end{table*}

\section{Results}
\label{sec:results}

\subsection{Context-length ablation}
Extending the text encoder context length consistently improves retrieval performance and training efficiency.
Table~\ref{tab:retrieval_all} shows the zero-shot image-to-text and text-to-image recall at $k$ in the long context benchmarks. 
Panel A shows that longer context improves retrieval across recall levels for both CXR and PMC.
On CXR, gains are steady but relatively modest. 
PMC benefits most, especially at stricter thresholds, highlighting the value of long contexts for text-heavy tasks.
In addition, we observe that pretraining CLIP with longer context lengths accelerates convergence (appendix section~\ref{apd:converge}), indicating that context extension improves not only downstream retrieval but also training efficiency.

% \begin{figure*}[htbp]
%   \centering
%  % \fbox{\rule{0pt}{200pt}\rule{150pt}{0pt}} % height=100pt, width=150pt
%   \includegraphics[width=0.8\linewidth]{}
%   \caption{
%   % (A) BIOMEDICA-6M caption token length distribution 
%   Zero-shot accuracy as a function of text context length. A) Zero-shot image classification. B) Zero-shot image-text and text-image retrievel.}
%   \label{fig: motivation-and-example}
% \end{figure*}

\subsection{Benchmarking against baselines}
BMC-LongCLIP outperforms prior biomedical VLMs on both long-text benchmarks. Table~\ref{tab:retrieval_all} Panel B and Table~\ref{tab:zeroshot} compare BMC-LongCLIP with existing biomedical VLMs, including PMC-CLIP, BiomedCLIP, MedSigLIP, and BMC-CLIP. We exclude MedSigLIP from the CXR benchmark comparison, as it was trained on the same MIMIC-CXR image–report pairs used to construct our benchmark.

% On the CXR benchmark, baselines achieve $<$6\% Recall@10, while BMC-LongCLIP variants reach 10–14\%, a $>$2× improvement. On PMC, BMC-LongCLIP reaches 89–95\% R@10, matching/surpassing BiomedCLIP and far outperforming MedSigLIP; gains are largest at stricter thresholds.

On the CXR benchmark, baselines achieve $<$6\% Recall@10, while BMC-LongCLIP variants reach 10–14\%, a more than two-fold improvement. 
On PMC, BMC-LongCLIP achieves 89–95\% R@10, performing on par with or slightly better than BiomedCLIP (91–94\%) and outperforming MedSigLIP (46–60\%). At the stricter R@1 threshold, BMC-LongCLIP attains 69–81\%, surpassing BiomedCLIP (69–73\%) and outperforming MedSigLIP (20–31\%).

% On PMC, BMC-LongCLIP attains 89–95\% R@10, matching or slightly surpassing BiomedCLIP (91–94\%) and outperforming MedSigLIP (46–60\%), with the largest relative gains at stricter thresholds. For R@1, BMC-LongCLIP reaches 69–81\%, exceeding both BiomedCLIP (64–72\%) and far surpassing MedSigLIP (21–36\%)

% On the CXR benchmark, baseline models perform poorly, with Recall@10 values below 6\%. In contrast, BMC-LongCLIP-512 variants achieve Recall@10 between 10.3--12.2\% (T2I) and 9.8--14.5\% (I2T). Across all BMC-LongCLIP variants, this represents more than a two-fold improvement over the strongest baseline.

% On the PMC benchmark, BMC-LongCLIP variants achieve Recall@10 scores ranging from 89.3\% to 95.1\% (T2I) and 89.9\% to 93.8\% (I2T). These results closely match or slightly surpass BiomedCLIP (91.1 and 93.7\%) and substantially outperform MedSigLIP (45.5 and 60.7\%). While we highlight R@10 for clarity, the same trend holds across all recall levels, with the largest relative gains observed at stricter thresholds such as R@1.

Beyond retrieval, BMC-LongCLIP also improves zero-shot classification accuracy across six biomedical domains (Table~\ref{tab:zeroshot}). 
While BiomedCLIP and MedSigLIP achieve average accuracies of 41.9\% and 36.6\%, respectively, BMC-LongCLIP (8K) attains 50.2\%, the best overall performance. 
These results highlight that extending context length not only benefits long-text retrieval but also provides improvements in classification tasks.

\subsection{Effect of batch size and long-caption training} Long-context models benefit from enriched captions, while larger batch sizes yield mixed results. Doubling batch size with BMC-LongCLIP (16K) underperforms in microscopy and dermatology. This suggests that long context windows combined with large batch sizes may not uniformly translate into performance gains across domains. In contrast, BMC-LongCLIP+ (16K) with BIOMEDICA-LongCAP data recovers this drop and matches or exceeds the 8K model. These results indicate that long-context modeling is most effective when paired with sufficient long-caption supervision, though performance in microscopy remains comparatively weak and warrants further investigation.
\noindent \textbf{Overall.} 
Across all experiments, BMC-LongCLIP outperforms prior biomedical VLMs in long-text retrieval and provides competitive advantages in zero-shot classification.
\section{Conclusion}
\label{sec:conclusion}

Our results show that extending text context length in biomedical VLMs delivers clear gains.
On long-text retrieval tasks, BMC-LongCLIP outperforms prior baselines on both CXR and PMC benchmarks, with the largest gains observed for PMC benchmark.
A key limitation is the scarcity of long-text benchmarks across biomedical domains; expanding such resources will be essential for a fuller evaluation of long-context models.
Taken together, these results establish long-context modeling as a promising direction for advancing biomedical VLMs.

\acks{This research was supported by grants from NVIDIA and utilized NVIDIA A100 GPUs. We gratefully acknowledge additional support from the AIMI–AWS Cloud Credit Program, which provided AWS cloud computing.

We further acknowledge support from the Stanford Data Science Scholars fellowship and ARPA-H to M.S., and the Arc Institute Graduate Fellowship to A.L. This work was also supported by NIH grant R01 GM134483 to R.T., the Hoffman-Yee Research Grant to S.Y.L., and NSF grant 19DMS1208164. S.Y.L. is a Chan Zuckerberg Biohub – San Francisco Investigator.}

%\newpage

\bibliography{jmlr-sample} 
% \bibliography{references} 

\appendix

% \section{Related Work}\label{apd:related}

\newpage
\section{BIOMEDICA-LongCAP details}\label{apd:longcap}

BIOMEDICA-LongCAP data generation pipeline using Qwen2-VL-72B-Instruct \cite{qwen2vl}:

\begin{enumerate}
    \item {\bf Context-Aware Caption Augmentation} We enhance the original figure caption by contextualizing the image description with additional information from the full-text article. Specifically, we collect the original caption, inline mentions from the main text, the abstract, and acronyms used throughout the aforementioned data. Then a VLM is prompted to augment the original caption, by only leveraging the provided information.
    
    \item {\bf Feasibility Assessment.} Given an image and its augmented caption, we extract all atomic features from the generated caption and prompt the VLM to evaluate whether it is feasible to discern each feature from the image alone—without relying on external sources or any information not visually present, unless explicitly overlaid with feasibility text. The output is an XML file in which each atomic feature is labeled as either \texttt{FEASIBLE} or \texttt{NOT\_FEASIBLE}, along with a rationale explaining the label.

    \item {\bf Caption Refinement via Feasibility Filtering.} Based on the feasibility assessment, we generate a refined caption that preserves only atomic features labeled as \texttt{FEASIBLE}. Features labeled as \texttt{NOT\_FEASIBLE} are removed or reworded to ensure that the final image description reflects only information that can be visually supported.
    
    \item {\bf Acronym Expansion.} While all previous steps had access to acronym definitions, we explicitly expand all acronyms based on a curated acronym list derived from the full-text article. This ensures that the captions are readable and unambiguous.
    
\end{enumerate}

% \section{CXR Benchmark}
% % histogram of token length
% We evaluate cross-modal retrieval between radiology images and their paired full-text reports. Each image--report pair is treated as one ground-truth match (1:1 setting). Reports are tokenized with the CLIP tokenizer.
% All embeddings are L2-normalized. Performance is reported as Recall@\{$1,5,10,100$\} for both text$\to$image and image$\to$text directions.

% \section{PMC Benchmark}
% We follow the same retrieval setup as in the CXR benchmark.
\section{CXR Benchmark}
We evaluate cross-modal retrieval between chest radiographs and their paired full-text reports. 
Let $\mathcal{D} = \{(x_i, t_i)\}_{i=1}^N$ denote the dataset, where $x_i$ is an image and $t_i$ its paired report. 
Each pair is treated as a one-to-one ground-truth match. 

Reports are tokenized with the CLIP tokenizer and embedded via the text encoder $E_{\text{text}}$, while images are preprocessed and embedded via the vision encoder $E_{\text{img}}$. 
We obtain
\[
z^{\text{text}}_i = \frac{E_{\text{text}}(t_i)}{\|E_{\text{text}}(t_i)\|}, \quad 
z^{\text{img}}_i = \frac{E_{\text{img}}(x_i)}{\|E_{\text{img}}(x_i)\|},
\]
where all embeddings are L2-normalized. 

For text$\to$image retrieval, we rank all image embeddings $\{z^{\text{img}}_j\}$ by cosine similarity with a query $z^{\text{text}}_i$; the ground-truth match is $z^{\text{img}}_i$. 
Image$\to$text retrieval is defined analogously. 
Performance is reported as Recall@$\{1,5,10,100\}$ for both directions.

We analyzed the token length distribution of the 1,000 reports in our evaluation set using the BioClinical-ModernBERT tokenizer. The reports contained on average 168.3 tokens, with a median of 158 tokens. The shortest report had 49 tokens, while the longest contained 427 tokens.

\section{PMC Benchmark}
We adopt the same retrieval formulation as CXR Benchmark on biomedical articles from PubMed Central. To construct the benchmark, we used the PubMed Central FTP service to download media bundles containing both .nxml full-text files and associated image files for recently published 2025 articles. From each article, we sampled exactly one unique image–caption pair and did not reuse articles across the benchmark, ensuring that each pair represents a distinct source document.

For the 1,000 CXR reports, tokenization with BioClinical-ModernBERT yielded an average length of 510 tokens, with a median of 460 tokens. Report lengths ranged from 251 tokens at the lower end to 1,022 tokens at the upper bound.

\section{Zero-shot classification benchmark}\label{apd:zeroshot}
For the detailed dataset provenance, including dataset names, citations, modalities, and class counts, please refer to BIOMEDICA \cite{biomedica}, Table S8. Each dataset’s classification task is reformulated into a closed-form VQA task. Labels are mapped to short human-readable text descriptions, and each image is paired with a multiple-choice list of candidate answers (including distractors). The correct label is randomly permuted among the options, and evaluation is performed by computing the similarity between image and answer embeddings.

% see table S8 in biomedica 

\section{Compute details}\label{apd:compute}

\begin{table}[h]
\centering
\begin{tabular}{l c c}
\toprule
\rowcolor{gray!10} \textbf{GPU Model} & \textbf{GPU Memory} & \textbf{Quantity} \\
\midrule
NVIDIA H200 & 141 GB & 8 \\
NVIDIA A100 & 80 GB & 16 \\
\bottomrule
\end{tabular}
\caption{Compute resources used for training.}
\end{table}

\section{Training parameters}\label{apd:param}

\begin{table}[h!]
\centering
\scriptsize
\setlength{\tabcolsep}{4pt} % reduce column spacing
\begin{tabular}{ll}
\toprule
\rowcolor{gray!10} \textbf{Model} & \textbf{Hyperparameters} \\
\midrule
\multirow{12}{*}{BMC-LongCLIP} 
 & context length: 77/154/512 \\ 
 & batch size (per GPU): 1024 \\ 
 & GPUs: 8$\times$H200 \\
 & effective batch size: 8192 \\
 & learning rate: $5\mathrm{e}{-4}$ \\ 
 & beta1: 0.9,\; beta2: 0.95 \\ 
 & warmup: 1000 \\ 
 & max epochs: 20 \\ 
 & precision: FP32 \\ 
 & grad. clip norm: 1.0 \\ 
 & dataset type: WebDataset \\
 & dataset: Biomedica-6M  \\ 
\cmidrule{1-2}
\multirow{12}{*}{BMC-LongCLIP+} 
 & context length: 512 \\
 & batch size (per GPU): 1024 \\ 
 & GPUs: 16$\times$A100 \\
 & effective batch size: 16384 \\
 & learning rate: $5\mathrm{e}{-4}$ \\ 
 & beta1: 0.9,\; beta2: 0.95 \\ 
 & warmup: 1000 \\ 
 & max epochs: 20 \\ 
 & precision: FP32 \\ 
 & grad. clip norm: 1.0 \\ 
 & dataset type: WebDataset \\
 & dataset: Biomedica-6M + LongCAP  \\ 
\midrule
\end{tabular}
\caption{Hyperparameters used for pretraining.}
\label{tab:hyperparam_search_grid}
\end{table}

\newpage

\section{Training loss curves by training context length}\label{apd:converge}
\begin{figure}[htbp]
  \centering
 % \fbox{\rule{0pt}{200pt}\rule{150pt}{0pt}} % height=100pt, width=150pt
  \includegraphics[width=1\linewidth]{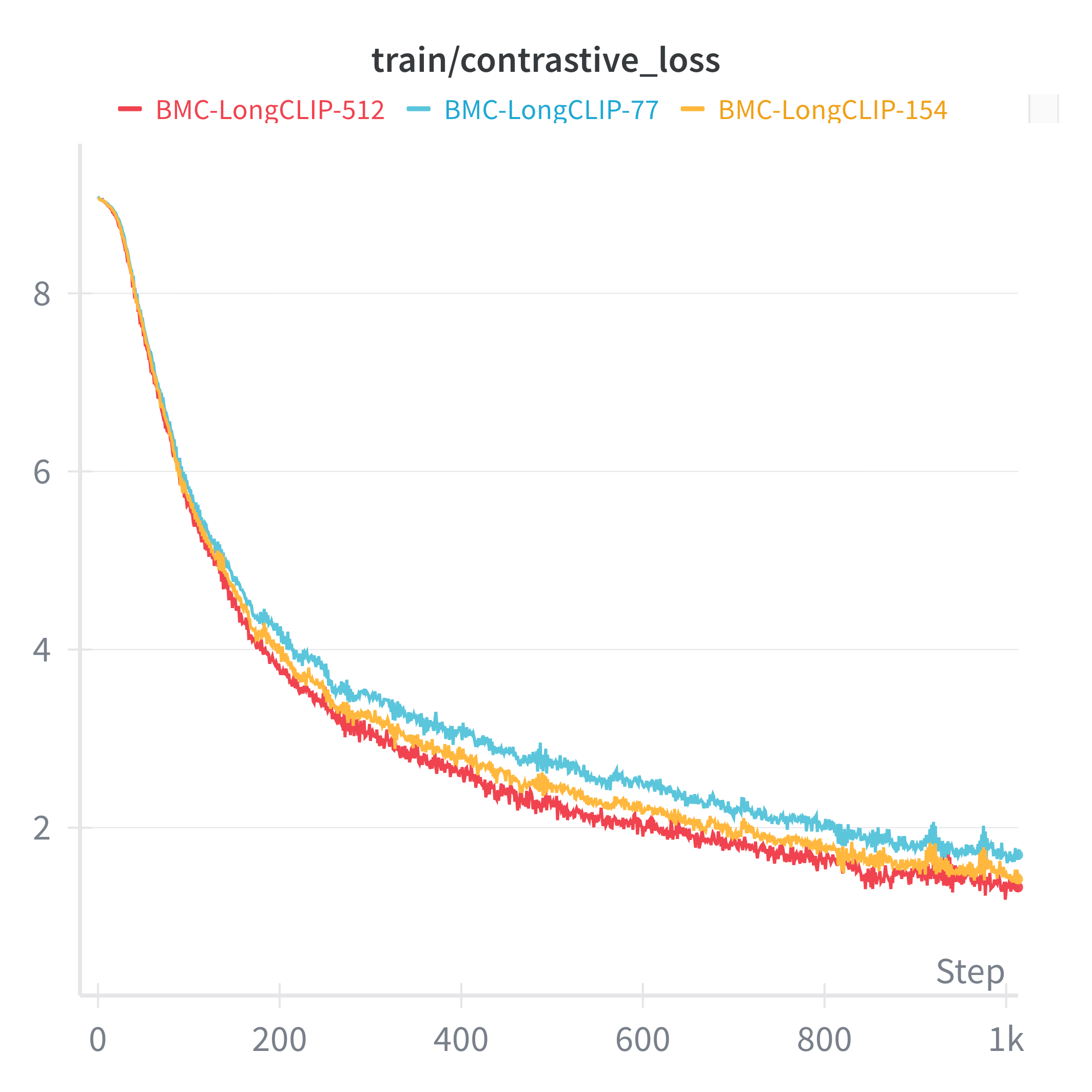}
  \caption{Training loss curves across context lengths, illustrating that longer text windows accelerate convergence.}

  \label{fig: convergence}
\end{figure}

\end{document}